\documentclass[conference]{IEEEtran}
\IEEEoverridecommandlockouts
\usepackage{cite}               
\usepackage{hyperref}           
\usepackage{url}                

\usepackage{amsmath,amssymb,amsfonts} 

\usepackage{graphicx}           
\usepackage{subcaption}         
\usepackage{float}              
\usepackage{caption}            

\usepackage{booktabs}           
\usepackage{array}              
\usepackage[table,xcdraw]{xcolor} 

\usepackage{algorithmic}        
\usepackage{listings}  
\usepackage{multirow}

\usepackage{todonotes}          
\usepackage{tcolorbox}

\usepackage{textcomp}           

\graphicspath{{media/}}

\def\BibTeX{{\rm B\kern-.05em{\sc i\kern-.025em b}\kern-.08em
    T\kern-.1667em\lower.7ex\hbox{E}\kern-.125emX}}
\begin{document}

\title{Towards LLM-based optimization compilers. Can LLMs learn how to apply a single peephole optimization? Reasoning is all LLMs need!\\
}

\author{\IEEEauthorblockN{1\textsuperscript{st} Xiangxin Fang}
\IEEEauthorblockA{\textit{Queen Mary University of London} \\
London, United Kingdom \\
jp2019213661@qmul.ac.uk}
\and
\IEEEauthorblockN{2\textsuperscript{nd} Lev Mukhanov}
\IEEEauthorblockA{\textit{Queen Mary University of London} \\
London, United Kingdom \\
l.mukhanov@qmul.ac.uk}
}

\maketitle
\pagestyle{plain}
\begin{abstract}
Large Language Models (LLMs) have demonstrated great potential in various language processing tasks, and recent studies have explored their application in compiler optimizations. However, all these studies focus on the conventional open-source LLMs, such as Llama2, which lack enhanced reasoning mechanisms. In this study, we investigate the errors produced by the fine-tuned 7B-parameter Llama2 model as it attempts to learn and apply a simple peephole optimization for the AArch64 assembly code. We provide an analysis of the errors produced by the LLM and compare it with state-of-the-art OpenAI models which implement advanced reasoning logic, including GPT-4o and GPT-o1 (preview). We demonstrate that OpenAI GPT-o1, despite not being fine-tuned, outperforms the fine-tuned Llama2 and GPT-4o.

Our findings indicate that this advantage is largely due to the chain-of-thought reasoning implemented in GPT-o1. We hope our work will inspire further research on using LLMs with enhanced reasoning mechanisms and chain-of-thought for code generation and optimization.
\end{abstract}

\begin{IEEEkeywords}
Compiler Optimization, Large Language Model, Reasoning
\end{IEEEkeywords}

\section{Introduction}
Large Language Models (LLMs) have made significant advancements in various software engineering tasks, showcasing their ability to understand and generate human-like code. Numerous studies have demonstrated the capability of these models to handle various high-level programming languages effectively. In code generation, tools like GitHub Copilot \cite{peng2023impactaideveloperproductivity} and OpenAI Codex \cite{chen2021evaluatinglargelanguagemodels} have proven their utility by assisting developers in creating relevant code snippets. In code translation, systems such as Facebook’s TransCoder \cite{lachaux2020unsupervisedtranslationprogramminglanguages} and Microsoft’s CodeBERT \cite{feng2020codebertpretrainedmodelprogramming} have shown effectiveness in translating code between different programming languages. 

The use of LLMs for compiler optimizations and code generation is a promising alternative to the classical compilers\cite{armengolestapé2022learningcx86translation}. LLMs can save significant effort and cost in developing new compilers, extending the existing native and binary compilers for new ISAs, supporting new programming languages, improving decompilers\cite{armengolestapé2024sladeportablesmalllanguage,Hosseini_2022} or even deriving new ISA-specific compiler optimizations. 
Recent studies have demonstrated that LLMs can be applied for decompilation\cite{armengolestapé2024sladeportablesmalllanguage,Hosseini_2022} and compiling the native C code into X86\cite{armengolestapé2022learningcx86translation}. Meta went even further by releasing foundation LLMs for code compilation and optimization\cite{cummins2024metalargelanguagemodel}. However, these studies focus on investigating the capabilities of conventional LLMs, such as Llama2, which lack of enhanced reasoning mechanisms, to optimize and generate the code. Moreover, these studies often tend to train on IRs or binaries where various combinations of compiler optimizations are applied. This prevents detailed analysis of the LLM errors, making it unclear to what extent these models can truly understand and optimize the code.

\textbf{The main goal} of our study is to investigate if advanced reasoning mechanisms, such as chain-of-thought, can help LLMs to improve their code generation and optimization capabilities. We want to understand when and what types of errors are produced by conventional LLMs and if reasoning can help to reduce these errors.

In this study, we focus specifically on peephole optimization since LLMs are sensitive to the context window size,
which hinders the optimization of code samples containing
many instructions\cite{cummins2023largelanguagemodelscompiler}. We exploit the fact that peephole examines and applies optimizations to each basic block where
the number of instructions is limited. Thus, we can investigate the strengths of LLMs in handling short instruction sequences, mitigating the limitations of LLMs in handling the context with a big size. We also specifically target peephole optimization since it is a fundamental part of any compiler, performing basic algebraic transformations. Thus, if LLMs cannot perform this optimization correctly, it is unlikely they will be able to handle more sophisticated optimizations.

\textbf{The contribution} of this study can be summarized as follows:
\begin{itemize}
\item we present the results of our experimental study on investigating the capabilities of the 7B-parameter Llama2 model \cite{touvron2023llamaopenefficientfoundation} to learn and apply a basic peephole optimization. Specifically, we fine-tune the model using 100,000 AArch64 basic block samples generated by the LLVM compiler. We investigate the syntactic and output correctness of the code samples optimized by the model using 24,000 basic blocks extracted from Code-Force, Code-Jam and BigCode datasets.  

\item We provide a detailed study and analysis of the errors produced by the Llama2 model when applying peephole optimization. We demonstrate that the induced errors are primarily due to a lack of reasoning. We make several important observations, such as the fact that fine-tuning tends to adjust the model to fully mimic the compiler's behavior which negatively affects the model's generalization capabilities.

\item We compare the accuracy of the fine-tuned Llama2 model with state-of-the-art OpenAI models, such as GPT-4o and GPT-o1 (preview). We demonstrate that GPT-o1 significantly outperforms the fine-tuned Llama2 and GPT-4o by enabling chain-of-thought.

Finally, we demonstrate that the number of steps in chain-of-thought and inference time have a strong impact on the performance of GPT-o1. 

\end{itemize}

The paper is organized as follows:  Section \ref{sec:background} presents background and our framework design; Section \ref{sec:eval_base} presents the evaluation results for our baseline model, i.e. the fine-tuned Llama2 model; Section \ref{sec:eval_reason} presents the evaluation results for the models with enhanced reasoning; Section \ref{section:limitations} discusses the limitations of our study and future research; Section \ref{sec:related_work} presents related work
and Section \ref{sec:conclusion} discusses the conclusion.
\section{Background and Framework design}
\label{sec:background}
\textbf{Compiler and ISA:} Peephole is an optimization technique applied at the basic block level that involves replacing the instructions with a logically equivalent set that has better performance \cite{muchnick1997advanced}. Peephole employs numerous pattern matching algorithms to apply a wide range of optimization scenarios. Its replacements include but are not limited to the following \footnote{We provide examples of typical optimizations performed by peephole in Table \ref{tab:optimization_exmpl} for the readers who are not familiar with compilers.}\cite{fischer2010crafting} :
\begin{itemize}
    \item Null sequences – Remove instructions which results are not used.
    \item Combine operations – Replace multiple operations with a single equivalent one.
    \item Algebraic optimizations – Apply algebraic laws to simplify or rearrange instructions.
    \item Address mode operations – Memory address algebraic optimizations.
    \item Special case instructions – Use of special instructions and operands, such as zero operand.
\end{itemize}

In our study, we generate AArch64 v8 code samples using the LLVM compiler \cite{LLVM:CGO04} (version 14.0.0) to train LLMs.

LLVM contains two default peephole implementations: \textit{Instcombine} \cite{llvm_instruction_combining} and \textit{Aggressive-instcombine} \cite{llvm_aarch64_peephole_opt}. The first implementation includes a large number of optimizations listed above, and all the transformations are applied within an optimized basic block \cite{llvm_instruction_combining} (algorithmic complexity is \textit{O(1)}). Meanwhile, the second implementation is more aggressive, i.e. algorithmic complexity is higher than \textit{O(1)}, and it can also modify CFG.

Given the mentioned limitation of the context size\cite{cummins2023largelanguagemodelscompiler}, we use \textit{Instcombine} in our study to train LLMs.

\begin{figure}[htbp]
\centering
\includegraphics{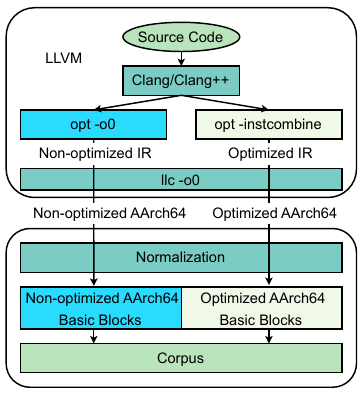}
\caption{Basic block generation pipeline.}
\label{fig3}
\end{figure}

\textbf{Generation of the training data:} To investigate whether LLMs can learn how to apply peephole, we use the LLVM compiler to generate optimized and non-optimized versions of basic blocks. Figure \ref{fig3} shows the code generation pipeline used in our study. We first compile the source code into IR using the \texttt{clang}/\texttt{clang++} frontend. The non-optimized IR is generated by invoking \texttt{opt} with the \texttt{-O0} flag. For generating the optimized IR, we use the \texttt{opt -instcombine} command to perform a series of peephole optimizations. These IRs are compiled into non-optimized and optimized AArch64 assembly code using \texttt{llc} with the \texttt{-O0} flag. Finally, we parse the generated AArch64 assembly codes and match optimized basic blocks with non-optimized ones. We also apply the following pre-processing steps (normalization):

\begin{itemize}
\item Remove metadata and attributes.
\item Remove basic blocks with more than 15 lines of instructions\footnote{We limit the size of basic blocks due to constraints of our computing facilities. Additionally, access to the OpenAI models used in our study is a paid service, with costs incurred based on the number of tokens processed.}.
\item Remove duplicate basic blocks with identical instructions.
\end{itemize}

We apply these pre-processing steps to follow previous studies\cite{cummins2023largelanguagemodelscompiler}.

\begin{figure*}[htbp]
\centering
\includegraphics[width=1\textwidth]{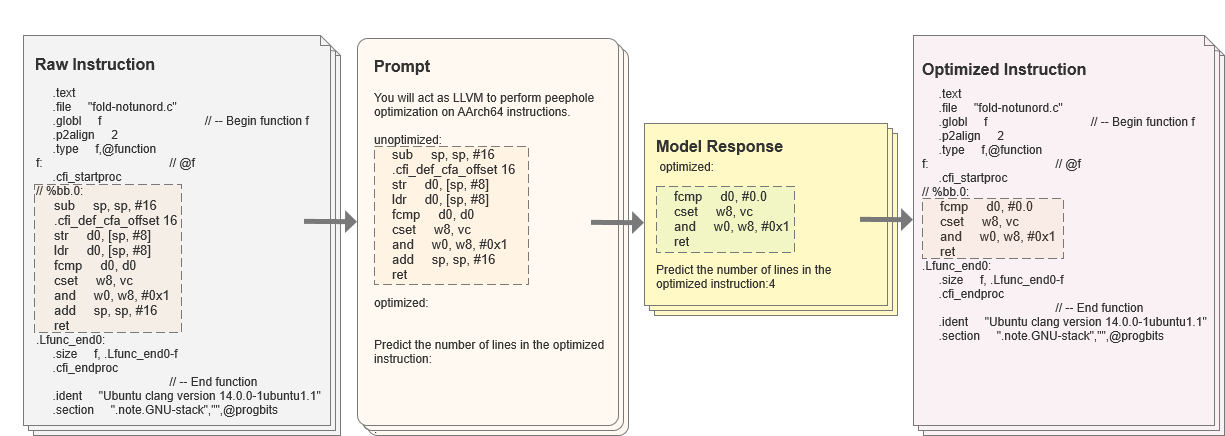}
\caption{The code optimization pipeline.}
\label{Model Flow}
\end{figure*}

\textbf{The LLM-based optimization pipeline:} Figure \ref{Model Flow} demonstrates the LLM-based code optimization pipeline. We first compile the source code and generate AArch64 assembly. Then we apply the pre-processing steps discussed previously to each basic block. 

We embed each basic block into a specific prompt template to guide LLMs in performing peephole optimization. Once a model generates the optimized basic block, we replace the original basic block with the optimized one and generate a binary\footnote{Note that there are limitations of such an approach which we discuss in Section \ref{section:limitations}.}. 

\textbf{LLMs:} In this study, we use the 7B-parameter Llama2 model as a baseline model, which is widely recognized for outperforming many other open-source models across multiple benchmarks \cite{touvron2023llama2openfoundation}. 

Apart from the Llama2 model, we use pre-trained OpenAI GPT-4o \cite{openai2024gpt4ocard} and GPT-o1 (preview) \cite{openai2024o1systemcard} models to investigate code optimization efficiency for LLMs with enhanced reasoning logic. The difference between these models is that GPT-o1 provides mode advanced structured reasoning and problem-solving capabilities\cite{touvron2023llama2openfoundation}.

\begin{table}[htbp]
\renewcommand{\arraystretch}{1.2}
\caption{Examples demonstrating the Llama2 model's understanding of AArch64 instructions (without fine-tuning) are shown. Cases (a) and (b) demonstrate that Llama2 accurately interprets the instructions, while case (c) illustrates a more complex scenario where Llama2 fails to correctly comprehend the instructions.}
\label{tab:LlaMa2_aarch64}
\normalsize
\begin{tabular}{p{8cm}}
\hline
\textbf{Prompt:} What does \texttt{ldr} mean in AArch64 instructions? \\
\textbf{Response:} The \texttt{ldr} instruction loads a 64-bit value from memory into a 64-bit register. \\
\textbf{(a)} This response is correct. \\
\hline
\textbf{Prompt:} What does \texttt{sub} mean in AArch64 instructions? \\
\textbf{Response:} A \texttt{sub} instruction in AArch64 is an instruction that performs a subtraction operation on two 64-bit operands. \\
\textbf{(b)} This response is correct. \\
\hline
\textbf{Prompt:} What does \texttt{sxtw} mean in AArch64 instructions? \\
\textbf{Response:} In AArch64 (64-bit ARM) instructions,\texttt{sxtw} is a mnemonic that stands for "shift extract with write" \\
\textbf{(c)} Incorrect. The \texttt{sxtw} stands for "Sign Extend Word." \\
\hline
\end{tabular}
\label{example_query}
\end{table}

\textbf{Fine-tuning of the baseline model:} 
Llama2 is pre-trained on a large corpus, with 4.5\% of the data sourced from public code on GitHub \cite{touvron2023llamaopenefficientfoundation}.
We directly queried the model to assess its understanding of the specific meanings of AArch64 instructions. Table \ref{example_query} presents three distinct questions related to AArch64 instructions. While the model demonstrates an ability to understand some instructions, it does not recognize more complex instructions.

To explore this further, we provided the model with several prompt templates (shot learning prompts) prior to submitting optimization requests to guide its understanding. However, our results indicate that the model was still unable to perform the requested optimizations. To enhance accuracy of the Llama2 model, we apply a fine-tuning process \cite{han2024parameterefficientfinetuninglargemodels}.

In our study, we also use pre-trained OpenAI GPT-4o and GPT-o1 models. However, we cannot fine-tune these models since we do not have access to source code.

\textbf{Quantization of the baseline model:} A complete fine-tuning of LLMs is very expensive. When using the regular AdamW optimizer \cite{loshchilov2019decoupledweightdecayregularization}, the Llama2-7B model requires a minimum of 8 bytes per parameter for its 7 billion parameters during training, totaling 56 GB of GPU memory\footnote{In our experimental framework, GPUs have only 40Gb of DRAM.}. To reduce memory usage, we employ QLoRA (Efficient Fine-tuning of Quantized LLMs) \cite{dettmers2023qloraefficientfinetuningquantized}, which significantly decreases the parameter storage requirements.

QLoRA applies the following major three techniques:
\begin{enumerate}
    \item {4-bit NormalFloat Quantization:} This reduces the precision of model parameters to 4 bits while maintaining numerical stability.
    \item {Dual Quantization:} This technique further reduces average memory usage by quantizing the quantization constants themselves.
    \item {Unified Memory Paging Technology\cite{nvidia2024cudacprogrammingguide}:}  This leverages NVIDIA's unified memory mechanism to enhance the efficiency of memory page transfers between CPU and GPU, thereby managing memory peak values. 
\end{enumerate}
We applied low-rank adaptation with a rank of 8 to the query and value projection layers of Llama2 and used the standard QLoRA configuration. This technique reduces the trainable parameter number by 1606x,  retaining approximately 0.06\% of the original parameters.

\textbf{Optimization of the baseline model:}
The model, with 4,194,304 trainable parameters, is trained using the Paged AdamW optimizer \cite{loshchilov2019decoupledweightdecayregularization} with a learning rate of $2*10^{-4}$ and a constant learning rate schedule. The total training batch size is 128, achieved through micro-batches of 32 and gradient accumulation steps of 4. We trained the model on an Intel-based Cloud server using NVidia A100 with 40Gb of internal DRAM (VRAM).

\section{Evaluation of the baseline model}
\label{sec:eval_base}
In this section, we present the evaluation results for our baseline model and compare our results with the previous studies. We aim to replicate the training and evaluation strategy used in a recently presented study \cite{cummins2023largelanguagemodelscompiler}.

\subsection{Model Evaluation Metrics}
To assess the performance of the baseline model, we use conservative natural language evaluation metrics. These metrics provide a quantitative measure of how closely the code optimized by LLMs aligns with the code optimized by the compiler. In addition to these metrics, we also use metrics to evaluate the syntactic correctness and the accuracy of the IO output for code generated by LLMs. 

\textbf{BLEU Score}: The BLEU (Bilingual Evaluation Understudy) score \cite{BLEU} is commonly used in machine translation to evaluate the similarity between machine-generated sentences and reference sentences. In our context, it measures how closely the basic blocks optimized by LLMs match the basic blocks optimized by the compiler.

\textbf{Exact Match Rate (EMR)}: This metric also enables us to compare the code samples generated by LLMs and code samples generated by the compiler. To be more specific, it measures the character-to-character complete matching rate for the basic blocks generated by LLMs and the basic blocks generated by the compiler. In other words, EMR only gives credit if the entire generated output exactly matches the reference output, making it a more binary metric than BLEU; an exact match implies a BLEU score of 1. 

\textbf{Syntactic Accuracy}: The syntactic accuracy measures the proportion of basic blocks generated by LLMs that can be correctly compiled without errors by LLVM. This metric ensures that the generated instructions are syntactically correct.

\textbf{IO Accuracy}: IO accuracy is used to check whether, for a given set of inputs, the code generated by LLMs produces the same output as the code optimized by the compiler. IO accuracy has been validated as an effective approach in prior research on code generation \cite{armengolestapé2022learningcx86translation}.

Importantly, we use these metrics to follow the  previous studies exploiting LLMs for code generation\cite{armengolestapé2022learningcx86translation, cummins2023largelanguagemodelscompiler}, enabling us to verify our results.
\subsection{Fine-tuning}
To tune the baseline model, we use C and C++ code samples from the LLVM and GCC test suites\cite{xiaowei2012compiler}. 

We specifically use these suites since both suites were designed to test the code generation 
 and optimization functionality of the compilers. Thus, we expect that these test suites ideally fit for the purposes of fine-tuning covering the majority of possible code optimization scenarios. 

We extracted 100,000 basic blocks (27 million tokens in total) from the test suites to fine-tune the Llama2 model. The code generation pipeline (see Figure \ref{fig3}) is applied to create both optimized and non-optimized basic block samples. 
We found that 6 epochs is optimal for our study, as increasing the number of epochs yields minimal changes in the target metrics, i.e. BLEU, EMR, Syntactic Accuracy and IO Accuracy. Note that the context window size is 512 tokens.

\subsection{Evaluation using the LLVM/GCC test suite}
Similar to the previous study \cite{cummins2023largelanguagemodelscompiler}, we first evaluate the model throughout the fine-tuning process. To be more specific, after every 10,000 samples used for fine-tuning we assess the model on a holdout validation set. This set consist of 5000 basic block samples that have not been used for training. 

\begin{figure}[htbp]
\centering
\includegraphics{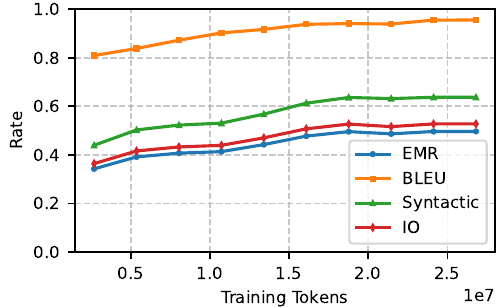}
\caption{Change in target LLM accuracy metrics with the number of tokens used for fine-tuning.}
\label{fig5}
\end{figure}

Figure~\ref{fig5} demonstrates how the target LLM accuracy metrics change if we increase the size of the data (the number of tokens) used for fine-tuning. We see that BLEU achieves a score of 0.92, indicating that the LLM generates the code which is quite similar to the code generated by LLVM. However, EMR achieves only 0.5 which implies that the model does not fully replicate the behavior of the compiler. Syntactic Accuracy and IO Accuracy reach 0.66 and 0.54, respectively. 

A previous study, which conducted similar experiments with the Llama2 model\cite{cummins2023largelanguagemodelscompiler}, has demonstrated results comparable to ours but slightly higher, i.e. BLEU at 0.952, EMR at 0.6 and Syntactic accuracy at 0.87. However, in this study, authors have trained LLMs to mimic the behavior of multiple compiler optimizations in order to reduce the size of the code. Moreover, authors trained the Llama2 model from scratch, while we performed fine-tuning. Finally, the previous study used much larger dataset, consisting of 1,000,000 IR functions and totaling approximately 10.9 billion tokens, compared to 27 million tokens in our study.

\subsection{Evaluation using Code-Force, Code-Jam and BigCode}
\label{sec:section3.4}

In this subsection, we present the results of the performance evaluation of the baseline model using a large dataset that combines thousands of algorithms implemented on major coding competition platforms. Our dataset is derived from two algorithmic competitions: Code-Force and Google Code Jam, with a particular focus on solutions written in C and C++. We have extracted the source code from GitHub applying pre-processing steps to remove duplicate code samples. In addition, we employed the BigCode's The Stack public repository, which is dedicated to training large language models with code\cite{lozhkov2024starcoder2stackv2}.

To extract basic blocks, we have processed about 300,000 C/C++ source code files, but we were able to compile and generate binaries only for 106,082 files. \footnote{The C/C++ files extracted from these datasets often include external dependencies, such as third-party libraries or project-specific headers. However, these header files and external libraries are not included in the datasets, which limits our ability to generate binaries. As a result, we were only able to obtain binaries for source files that use standard libraries.}
The detailed information about the number of extracted source code files and basic blocks is provided in Table \ref{tab:test_data}. Overall, we have managed to extract 437,892 basic blocks for which we can generate the binaries. Due to limitations in our computing resources, we randomly select 24,000 basic blocks to evaluate the performance of the fine-tuned Llama2 model.

\begin{table}[htbp]
\renewcommand{\arraystretch}{1.2}
\caption{Testing dataset info.}
\label{tab:test_data}
\centering
\small
\begin{tabular}{p{0.12\textwidth} p{0.07\textwidth} p{0.07\textwidth} p{0.07\textwidth}
p{0.07\textwidth}}
\toprule
  & \textbf{Source Files} & \textbf{Basic Blocks} & \textbf{Sampled Blocks} \\
\midrule
\textbf{Code-Force}  &1,050 & 5,610 & 4,000 \\
\textbf{Code-Jam}  & 10,764 & 110,354 & 10,000 \\
\textbf{BigCode} & 94,268 & 321,928 & 10,000 \\
\midrule
\textbf{Total} & 106,082 & 437,892 & 24,000 \\
\bottomrule
\end{tabular}
\normalsize
\end{table}

\subsection{Results}
\begin{figure}[htbp]
\centering
\includegraphics{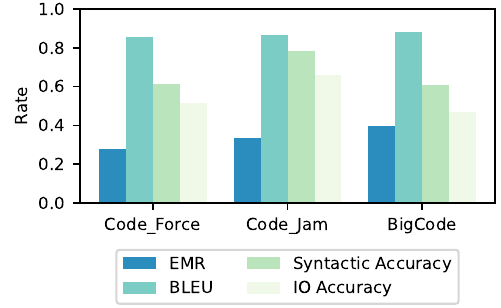}
\caption{Model performance on different test sets.}
\label{fig6}
\end{figure}

Figure~\ref{fig6} demonstrates BLEU, EMR, Syntactic accuracy and IO accuracy measured for 24,000 basic blocks optimized by the fine-tuned Llama2 model. Our first observation is that BLEU is almost the same for all the datasets, about 0.87, and it is slightly lower than BLEU measured in our previous experiments. However, EMR drops significantly compared to the previous experiments, and it varies from 0.27 to 0.39. This is explained by the fact that the training set significantly differs from the large testing set, and, as a result, the LLM cannot predict accurately the output generated by the compiler. However, we see only slight degradation of Syntactic Accuracy and IO accuracy drops compared to the previous experiments. These results suggest that the LLM has some generalization capabilities, as EMR decreases while the quality of the generated code, in terms of Syntactic Accuracy and IO Accuracy, remains almost the same.

Importantly, these results are aligned with the results obtained in the previous studies\cite{armengolestapé2022learningcx86translation,cummins2023largelanguagemodelscompiler}. For example, it is shown that the model translating C code into x86 instructions achieves a BLEU score of 0.79, a Syntactic Accuracy of 0.58, and an IO Accuracy of 0.33\cite{armengolestapé2022learningcx86translation}. 

\subsection{Error analysis}
Next, we conduct a systematic analysis of errors produced by the LLM. The model produced a total of 8,922 compilation errors. Note that we focus specifically on compilation errors, as they are easier to identify the sources of. We randomly select 1,000 of these errors to make a systematic analysis. Table \ref{tab5}a shows the distribution of 1,000 compilation errors across four different error types which we discuss below.

\textbf{Opcode errors} refer to invalid instruction codes, invalid instruction names, or inappropriate instruction names for a particular use generated by the LLM. An example of such errors is provided in Table \ref{tab6}. In this particular case, the LLM generated \textit{and} instruction instead of \textit{mov} instruction. Note that this type of error occurs most frequently in our experiments (see Table \ref{tab5}a).

\textbf{Immediate Value Errors} are related to the errors introduced by the LLM when using literal constants. 

For example, Table \ref{tab6} shows incorrect 
usage of literal constants. To be more specific, \textit{mov} instruction should initialize register \textit{w0} with 0 using register \textit{wzr}. However, it generates \textit{\#r} instead of \textit{wzr}. Note that \textit{\#} denotes an immediate value, and \textit{r} should be a literal constant in this particular case.

\textbf{Label Errors} represent the errors due to 
 incorrect labels and function names in assembly code. For example, in Table \ref{tab6}, \textit{adpr} instruction should load the address of \textit{.L.str} label. However, the LLM generated an incorrect label, i.e. \textit{ .Lstrstr}.

\begin{figure}[htbp]
\centering
\includegraphics{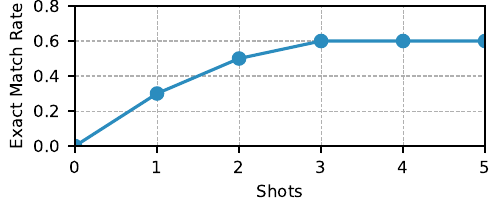}
\caption{Correlation between EMR (Exact Match Rate) and the number of prompt shots.}
\label{few-shot}
\end{figure}

\textbf{Register Errors} represent the errors induced by the LLM due to incorrect usage of registers. For example, in Table \ref{tab6} \textit{mov} instruction uses 32-bit register \textit{w8} instead of 64-bit register \textit{x8}.

Next, we further investigate the errors that occurred most frequently in our experiments, i.e. opcode errors. We estimated the probability of opcode errors for each instruction obtained in our testing dataset. Table \ref{tab5}b and \ref{tab5}c show top 10 instructions with the highest and lowest error probabilities, including 95\% confidence intervals, respectively. Note that these tables contain only those instructions which have a representative, i.e. a statistically significant, number of samples in the training dataset, i.e. above 50.

We observe that the error probability varies from 0.5\% for the \textit{stp} instruction to 43\% for the \textit{eor} instruction.The list of most error-prone errors include:

\begin{itemize}
\item \textit{eor (Bitwise exclusive-OR)}. The LLM often confuses this instruction with \textit{sub}, using it in place of \textit{eor}. It also confuses \textit{eor} with \textit{and}, \textit{mov} and \textit{ls};
\item \textit{ldurb} (Load Register byte). The model confuses unsigned byte loads (\textit{ldurb}) with signed loads, such as \textit{ldrsb} (Load Register Signed Byte) and \textit{ldrsh} (Load Register Signed Half-word);
\item \textit{asr} (Arithmetic Shift Right). The LLM tends to confuse this instruction with the logical shift instruction. Additionally, it often generates invalid instruction names, such as \textit{movr} and \textit{as+sr}.
\item \textit{adds}. The LLM struggles to differentiate between add instructions with and without flag setting, i.e. (\textit{add}) and (\textit{adds}). It also occasionally generates incorrect instructions, such as \textit{sub}, \textit{strs}, \textit{subs} and \textit{mov} instructions.
\item \textit{lsr}. The model often generates the non-existent \textit{movr} instruction, which is not part of Aarch64 ISA.
\item \textit{stur}. LLM generates various non-existent instructions, such as \textit{strur}, \textit{blur}, \textit{tbur}, \textit{movur}.
\end{itemize}
Our results indicate that, in the vast majority of error cases, the LLM model generated non-existent instruction names. 

One possible explanation for this could be  the lack of sufficient samples for certain instructions in the training dataset. However, even though \textit{stur} is one of the most frequently sampled instructions in the dataset (with 21,277 samples, see Table \ref{tab:train_data_stat}), it still ranks among the top 10 instructions with the highest error probability.

Interestingly, the \textit{stp} instruction has the lowest error probability (see Table \ref{tab5}c). However, in a few instances, the LLM generates non-existent \textit{subp} instruction instead of \textit{stp}. We attribute this to the frequent occurance of both \textit{stp} and \textit{sub} in the beginning of functions, leading the model to confuse these instructions and generate \textit{subp}. Such errors might be avoided if the LLM was guided by a reasoning-based approach rather than relying solely on the context.

\begin{tcolorbox}[boxrule=0pt,sharp corners, left=0pt, right=0pt, top=0pt, bottom=0pt, width=\columnwidth]
\textbf{Fundamental limitation of conventional LLMs for code generation.} Overall, our findings indicate that no instructions have an error probability of zero\footnote{The number of samples for which is representative, i.e. larger than 50.}, and the vast majority of errors are due to incorrectly generated instruction names. In other words, conventional LLMs exhibit some probability of error when generating opcodes for every instruction, and this probability is always greater than zero. We attribute this to the probabilistic nature of LLMs' output, which significantly limits the use of conventional LLMs for code generation. All the observations and errors, especially opcode errors and register errors, presented above highlight the need for LLMs to incorporate a reasoning mechanism and enhanced logic capabilities to overcome this fundamental limitation.
\end{tcolorbox}

\begin{table}[htbp]
\renewcommand{\arraystretch}{1.2} 
\raggedright
\caption{Distribution of compilation errors for Code-Force, Code-Jam and BigCore datasets.}
\label{tab5}
\small 

\begin{minipage}{\linewidth}
\centering
\subcaption{Distribution of 1000 compilation errors across 4 different types of errors. }
\begin{tabular}{lc}
\toprule
\textbf{Error Category} & \textbf{Count} \\
\midrule
Opcode Error & 526 \\
Immediate Value Error & 518 \\
Label Error & 272 \\
Register Error & 188 \\
\midrule
\textbf{Total} & \textbf{1000*} \\
\bottomrule
\end{tabular}
\vspace{0.5em}

\footnotesize{*The total number of errors does not match 1000 errors since some code samples have multiple errors.}
\end{minipage}

\vspace{1em}

\begin{minipage}{\linewidth}
\raggedleft
\subcaption{Opcode errors: Top 10 instruction types with the highest error probability; the number of samples $>$ 50.}
\begin{tabular}{lcccc}
\toprule
\textbf{Instr} & \textbf{Error Count} & \textbf{Total Count} & \textbf{Error Prob} & \textbf{Conf} \\
\midrule
eor     & 22    & 51    & 0.431373 & 0.135929 \\
ldurb   & 73    & 175   & 0.417143 & 0.073057 \\
asr     & 29    & 81    & 0.358025 & 0.104407 \\
adds    & 22    & 65    & 0.338462 & 0.115035 \\
lsr     & 24    & 72    & 0.333333 & 0.108889 \\
fmov    & 58    & 222   & 0.261261 & 0.057791 \\
stur    & 1381  & 5718  & 0.241518 & 0.011094 \\
tbnz    & 140   & 588   & 0.238095 & 0.034427 \\
ldursw  & 35    & 151   & 0.231788 & 0.067306 \\
orr     & 16    & 71    & 0.225352 & 0.097187 \\
\bottomrule
\end{tabular}
\end{minipage}

\vspace{1em}

\begin{minipage}{\linewidth}
\raggedleft
\subcaption{Opcode errors: Top 10 instruction types with the lowest error probability; the number of samples $>$ 50.}
\begin{tabular}{lcccc}
\toprule
\textbf{Instr} & \textbf{Error Count} & \textbf{Total Count} & \textbf{Error Prob} &\textbf{Conf}\\
\midrule
stp    & 37   & 7250   & 0.005103 & 0.001640 \\
ldp    & 37   & 5164   & 0.007165 & 0.002300 \\
ret    & 73   & 7028   & 0.010387 & 0.002370 \\
fadd   & 9    & 682    & 0.013196 & 0.008565 \\
bl     & 218  & 15087  & 0.014450 & 0.001904 \\
cbz    & 27   & 1145   & 0.023581 & 0.008789 \\
b      & 499  & 18584  & 0.026851 & 0.002324 \\
mul    & 37   & 1040   & 0.035577 & 0.011258 \\
sdiv   & 46   & 1192   & 0.038591 & 0.010935 \\
fcmp   & 7    & 181    & 0.038674 & 0.028091 \\
\bottomrule
\end{tabular}
\end{minipage}
\end{table}

\begin{table}
\centering 
\caption{Top 20 instructions with the most samples in the training dataset.}
\small
\begin{tabular}{l r @{\hspace{1cm}} l r @{\hspace{1cm}} l r}
\toprule
\cmidrule(lr){1-2} \cmidrule(lr){3-4} \cmidrule(lr){5-6}
\textbf{Instr} & \textbf{Count} & \textbf{Instr} & \textbf{Count} & \textbf{Instr} & \textbf{Count} \\
\midrule
b      & 107630 & subs    & 28560  & cbz    & 6880  \\
ldr    & 78803  & stur    & 21277  & mul    & 6231  \\
str    & 60305  & adrp    & 12884  & cbnz   & 5616  \\
mov    & 46248  & sub     & 11865  & ldp    & 5152  \\
add    & 42832  & ret     & 10814  & and    & 4553  \\
ldur   & 39714  & stp     & 9103   & strb   & 4051  \\
bl     & 30585  & udiv    & 8420   &        &        \\
\bottomrule
\end{tabular}
\label{tab:train_data_stat}
\end{table}

\begin{table*}[htbp]
\centering
\caption{Examples of errors introduced by Llama2.}
\label{tab6}
\begin{tabular}{lp{6.5cm}p{6.5cm}}
\toprule
\textbf{Error Type} & \textbf{Incorrect transformation} & \textbf{Correct transformation} \\
\midrule
Opcode Error & \textcolor{red}{mov} w8, w0, \#0xff\textbackslash n lsr w8, w8, \#4\textbackslash n orr w0, w8, w9 & \textcolor{red}{and} w8, w0, \#0xff\textbackslash n lsr w8, w8, \#4\textbackslash n orr w0, w8, w9 \\
Immediate Value Error & mov w0, \#\textcolor{red}{r}\textbackslash n ret & mov w0, \textcolor{red}{wzr}\textbackslash n ret \\
Label Error & adrp x0, \textcolor{red}{.Lstrstr}\textbackslash{}n add x0, x0, :lo12:.L.str & adrp x0, \textcolor{red}{.L.str}\textbackslash{}n add x0, x0, :lo12:.L.str \\
Register Error & mov \textcolor{red}{w8}, x0\textbackslash n mov w0, \#3\textbackslash n str w0, [x8]\textbackslash n ret & mov \textcolor{red}{x8}, x0\textbackslash n mov w0, \#3\textbackslash n str w0, [x8]\textbackslash n ret \\

\bottomrule
\end{tabular}
\end{table*}

\section{Evaluation of LLMs with advanced reasoning mechanisms}
\label{sec:eval_reason}
To investigate if LLMs can increase the rate of correct transformation by enabling reasoning mechanisms, we tested OpenAI GPT-4o and GPT-o1 (preview) which have advanced reasoning capabilities\cite{openai2024o1systemcard}.

\begin{table*}[ht]
\caption{The performance of LLMs.}
\centering
\centering
\begin{tabular}{llccc}
\toprule
\textbf{Test Samples} & \textbf{Metrics(\%)} & \textbf{Llama2 7B(Fine-tuned)} & \textbf{GPT-4o (3-shots)} & \textbf{GPT-o1 (3-shots)} \\
\midrule
\multirow{3}{*}{\textbf{ALL}} & EMR (Exact Match Rate) & 36.8 & 23.1 & -  \\
& Syntactic Accuracy & 64.0 & 86.6 & -  \\
& IO Accuracy & 51.6 & 54.2 & -  \\
& BLEU & 86.6 & 56.9 & -  \\
\midrule
\multirow{3}{*}{\textbf{Selected}} & EMR (Exact Match Rate) & 5.0 & 5.0 & 19.0 \\
& Syntactic Accuracy & 12.5 & 68.0 & 92.0 \\
& IO Accuracy & 7.3 & 49.1 & 79.1 \\
& BLEU & 73.0 & 35.0 & 78.0  \\
\bottomrule
\label{llms_results}
\end{tabular}
 
\label{performance}
\end{table*}

\subsection{The fine-tuned Llama2 vs GPT-4o}
We first evaluated OpenAI's GPT-4o model using our Code-Force, Code-Jam and BigCode datasets, as discussed in Section~\ref{sec:section3.4}. Since we cannot fine-tune GPT-4o (the source code is not available), we use 3-shot context learning prompts \footnote{3-shot learning implies that before requesting the models to optimize code, these models are provided 3 examples of the optimized and non-optimized code samples.} before requesting the model to optimize the code following the same pipeline as for the Llama2 model, see Figure~\ref{Model Flow}. Note that for all the optimization requests we use absolutely the same 3 learning prompts.

To motivate the use of prompt-based training, we conducted a series of experiments optimizing 20 randomly chosen basic blocks. We also selected 5 pairs of optimized and non-optimized basic blocks, which are randomly sampled, and use them as prompts. Figure \ref{few-shot} shows how the average EMR (the exact match rate for basic blocks generated by the LLM and blocks optimized by the compiler), measured across 20 basic blocks, changes with the number of prompts. We see that EMR grows more than 12x after applying the 3-shot prompt learning, and it reaches EMR about 60\%.

Note that EMR does not change when the number of shot prompts has increased above 3. Based on these results, we use the 3-shot prompt learning to evaluate GPT-4o using 24,000 basic blocks extracted from Code-Force, Code-Jam and BigCode datasets.

\begin{figure}[htbp]
\centering
\includegraphics{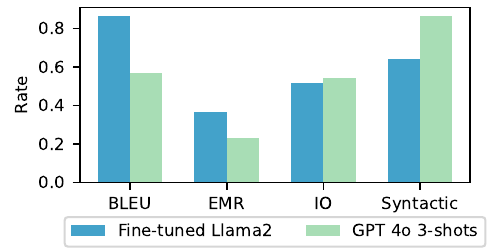}
\caption{Average BLEU, EMR, IO accuracy, Syntactic accuracy obtained for 24,000 basic block samples (Code-Force, Code-Jam and BigCode datasets).}
\label{GPT4o}
\end{figure}

We compare the fine-tuned Llama2 and GPT-4o using four key metrics: BLEU, EMR, IO accuracy, and Syntactic accuracy. Figure~\ref{GPT4o} shows the evaluation results. We see that the fine-tuned Llama2 outperforms GPT-4o in terms of BLEU and EMR metrics. This can be explained by the fact that the Llama2 model was specifically fine-tuned on peephole optimization code samples, aligning the code generated by the model with that produced by the specific compiler used in our study. Nonetheless, GPT-4o demonstrates a higher Syntactic accuracy and IO accuracy in our experiments which implies that this model has better capabilities to generate syntactically and semantically correct code. We attribute this to the fact that GPT-4o has a larger number of parameters and was trained on a diverse data. 

\begin{tcolorbox}[boxrule=0pt,sharp corners,left=0pt,right=0pt,top=0pt,bottom=0pt]
\textbf{Finding:} Based on these results, we conclude that BLEU and EMR metrics widely used in NLP \cite{vaswani2023attentionneed, brown2020languagemodelsfewshotlearners}studies do not really serve the purposes of code quality evaluation. Thus, the use of these metrics should be reconsidered in future research studies on code generation.
\end{tcolorbox}

\subsection{Llama2 vs GPT-4o vs GPT-o1}
In our final experiments, we tested the performance of an OpenAI model with the most advanced reasoning capabilities, i.e. GPT-o1 (preview)\footnote{https://openai.com/index/introducing-openai-o1-preview/}. Unfortunately, GPT-o1 is a paid model, which limits the number of code samples we could test in our study. To address this, we selected 200 most complicated code samples from 24,000 samples (Code-Force, Code-Jam and BigCode datasets) used for testing the fine-tuned Llama2 and GPT-4o. Note that we selected these samples in a such a way that the fine-tuned Llama2 model generates the code which differs from the the code produced by LLVM and produce errors in 190 samples; 
and we chosen 10 samples where both Llama2 and GPT-4o generate the correct code which is similar to the code generated by LLVM (200 samples in total). 
Such a division of the samples enables us to test GPT-o1 with both types of block samples, those correctly and those incorrectly optimized by Llama2. 

Table~\ref{llms_results} shows the evaluation results for all models tested with 24,000 samples (\textit{ALL}) and 200 selected samples (\textit{Selected}). 

 We see that GPT-o1 outperforms both GPT-4o and the fine-tuned Llama2 across three metrics for the 200 selected samples, i.e. EMR, IO accuracy, Syntactic accuracy. The fine-tuned Llama2 model shows the lowest Syntactic accuracy and IO accuracy for the selected samples. We believe this is due to the model's lack of generalization capabilities, as we found that if the Llama2 model generates code that differs from the one produced by LLVM, it is highly likely to result in a compilation error, program crash, or incorrect output. Meanwhile, GPT-4o and GPT-o1, which were not specifically fine-tuned on code samples with applied peephole optimization, often generate the correct code, although it does not match the code generated by LLVM. 

\begin{tcolorbox}[boxrule=0pt,sharp corners,left=0pt,right=0pt,top=0pt,bottom=0pt]
\textbf{Finding:} Based on these observations, we conclude that fine-tuning tends to adjust LLMs to completely mimic the behavior of compilers, which may negatively affect their generalization capabilities and overall performance.
\end{tcolorbox}

Importantly, although both GPT-4o and GPT-o1 have not been fine-tuned, GPT-o1 significantly outperforms GPT-4o. To be more specific, GPT-o1 achieves 14\% higher EMR, 24\% higher Syntactic Accuracy, and 30\% higher IO Accuracy compared to GPT-4o. We attribute such a difference to the enhanced reasoning capabilities of GPT-o1. 

To explore this, we made a deep investigation of a particular code sample for which both GPT-4o and the Llama2 model generate incorrect code, while GPT-o1 successfully produce the correct code. Table~\ref{tab:code-comparison} shows the original code sample and code samples optimized by LLVM, Llama2, GPT-4o and GPT-o1. The original code sample stores the value 5 at a 64-bit address contained in the \textit{x0} register with an offset based on the value in the 32-bit \textit{w1} register, shifted left by 2 and extended to 64 bits. However, this code contains several redundant instructions, including instructions writing and reading from the stack.

We see that GPT-4o made several errors. Specifically, it included a redundant move instruction (\textit{mov w8, w1}) and generated an incorrect load instruction which reads data from a stack address (\textit{[sp, \#12]}), while the model removed the store instruction which writes to this address. Thus, GPT-4o does not have enough reasoning capabilities to understand the optimization logic and avoid such simple errors. Nonetheless, the model generated syntactically correct code which can be compiled. Meanwhile, the Llama2 model generated syntactically incorrect code. Specifically, it generated incorrect opcodes, such as \textit{movsl} and \textit{movr}; the second argument for \textit{str w8, \#5} instruction also does not follow the assembly syntax. We explain this, as discussed previously, by the fact that fine-tuning adjust the model to mimic the behavior of the compiler which negatively affects the model generalization ability, especially for those code samples which significantly differ from samples used for fine-tuning. 

Finally, GPT-o1 correctly optimized the code. Moreover, we see that GPT-o1 generated better code than LLVM since it used only three instructions, while LLVM used five instructions. In particular, both GPT-o1 and LLVM removed instructions which allocate space in stack and use it which are redundant in this case. However, LLVM generated redundant \textit{lsl} and \textit{asr} instructions, while GPT-o1 performed the offset transformation by adding \textit{sxtw \#2} to \textit{str} instruction. Thus, we see that GPT-o1, with enhanced reasoning mechanisms, can outperform not only LLMs lacking advanced reasoning capabilities but also the compilers
\footnote{Note that the basic block represents an entire function in this case. Thus, there are no other basic blocks in this function which contain uses in the data-flow graph for the removed instructions. We also verified that the generated code executes correctly.}. 

\textbf{Reasoning and chain-of-thought}. 
We believe that the high performance of GPT-o1 compared to Llama2 and GPT-4o is primarily due to the implemented chain-of-thought mechanism. The key advantage of this mechanism is that the model makes several prompts to optimize the code in several stages, replicating the reasoning process. For example, Figure~\ref{chain_of_thought} shows the chain-of-thought applied by GPT-o1 to optimize the presented above code sample. We see that it is trying to understand the overall purpose of the provided code sample and examine the function flow. Moreover, it also builds a pseudo code for the provided code sample which is further used to generate the optimal code.

\begin{figure}[htbp]
\centering
\includegraphics{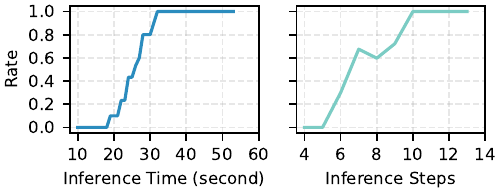}
\caption{The rate at which GPT-o1 produces the most optimal code, inference time and the number of inference steps.}
\label{o1_result}
\end{figure}

\textbf{But why is it specifically due to the chain-of-thought process?} Indeed, there are rumors that GPT-o1 implements more parameters than other models, which might allow it to outperform others due its size rather than chain-of-thought. To investigate this, we made several manipulations to vary time and the number of inference steps within the chain-of-thought process for a particular basic block optimization presented in Table \ref{tab:code-comparison}. We made 40 experiments adjusting the number of inference steps and inference time, and measured the rate at which GPT-o1 produces the most optimal code, i.e. the code presented in Table \ref{tab:code-comparison}e. Figure \ref{o1_result} illustrates how this rate changes with the number of inference steps and inference time. The results clearly show that the rate grows with both inference time and the number of inference steps. Notably, GPT-o1 generates the most optimal code in 100\% cases when the inference time exceeds 34 seconds or the number of inference steps is greater than 10 steps. In contrast, GPT-o1 never generates the optimal code when inference time is under 17 seconds or the number of steps is below 5. Furthermore, it also produces syntactically incorrect code in these cases. These findings suggest that the chain-of-thought mechanism significantly impacts GPT-o1's performance and the quality of the generated code.

Overall, we see that the chain-of-thought mechanism implemented by GPT-o1 significantly improves the performance of the model in generating optimized syntactically and functionally correct code. We believe that this is due to the fact that the model tries to understand the code and apply optimizations systematically. However, a better understanding the performance of LLMs with enhanced reasoning mechanisms for code generation and optimization requires a further research with open-sourced models.

\lstdefinelanguage
   [ARM]{Assembler}     
   {morekeywords={       
        add, sub, mov, str, ldr, ldrsw, ret, asr, lsl
    },
    sensitive=true,      
    morecomment=[l]{//}, 
    morestring=[b]",     
}

\lstset{
  basicstyle=\ttfamily\footnotesize,
  columns=fullflexible,
  frame=bottomline,
  breaklines=true,
  postbreak=\mbox{\textcolor{red}{$\hookrightarrow$}\space},
  language=[ARM]Assembler,
  keywordstyle=\color{blue},
  commentstyle=\color{green!50!black},
}

\begin{table}[htbp]
  \centering
  \caption{An example of code sample and peephole transformations performed by LLVM, Llama2, GPT-4o and GPT-o1.}
  \label{tab:code-comparison}

  \begin{tabular}{@{}p{\columnwidth}@{}}
  \subcaption{Original Code}
  \begin{lstlisting}
  sub sp, sp, #16
  .cfi_def_cfa_offset 16
  mov w8, w1
  str w8, [sp, #12]
  str x0, [sp]
  ldr x9, [sp]              // Load x9 with value at [sp] (x0)
  ldrsw x10, [sp, #12]      // Load signed word from [sp, #12] into x10, sign-extended to 64 bits
  mov w8, #5                // Move immediate 5 into w8
  str w8, [x9, x10, lsl #2] // Store w8 at [x9 + (x10 << 2)]
  add sp, sp, #16
  ret
  \end{lstlisting}
  \end{tabular}

  \begin{tabular}{@{}p{\columnwidth}@{}}
  \subcaption{Peephole performed by LLVM}
  \begin{lstlisting}
  lsl x8, x1, #32               // Logical shift left x1 by 32 bits, result in x8
  asr x9, x8, #32               // Arithmetic shift right x8 by 32 bits, sign-extend into x9
  mov w8, #5                    // Move immediate 5 into w8
  str w8, [x0, x9, lsl #2]
  ret
  \end{lstlisting}
  \end{tabular}

  \begin{tabular}{@{}p{0.48\columnwidth}@{\hspace{0.04\columnwidth}}p{0.48\columnwidth}@{}}

  \subcaption{GPT-4o Output} & 
  \subcaption{Llama2 Output} \\

  \begin{lstlisting}
  mov w8, w1
  mov w9, x0
  ldrsw x10, [sp, #12]      // Attempts to load from [sp, #12], but no valid data there
  mov w8, #5
  str w8, [x9, x10, lsl #2]
  ret
  \end{lstlisting}
  &
  \begin{lstlisting}
  movsl x8, x0, #2           // Invalid opcode 'movsl'
  movr x8, x0, #2            // Invalid opcode 'movr'
  str w8, #5                 // Incorrect syntax; 'str' requires a memory address
  str w8, [x0, x9, lsl #2]
  ret
  \end{lstlisting}

  \end{tabular}

  \begin{tabular}{@{}p{\columnwidth}@{}}
  \subcaption{GPT-o1 Output}
  \begin{lstlisting}
  mov w8, #5
  str w8, [x0, w1, sxtw #2]  // Store w8 at [x0 + (SignExtend64(w1) << 2)]
  ret
  \end{lstlisting}
  \end{tabular}

\end{table}

\begin{figure}[htbp]
\centering
\includegraphics[width=\columnwidth]{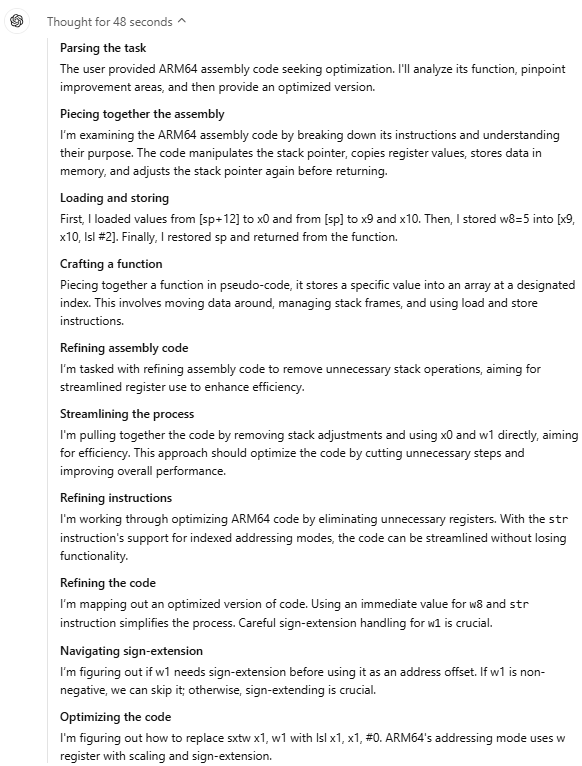}
\caption{A chain-of-thought produced by GPT-o1.}
\label{chain_of_thought}
\end{figure}
\section{Limitations and future research.}
\label{section:limitations}
There are several limitations in our study which we believe are important to discuss.

\textbf{Def-use graph across different basic blocks.} 
In our study, LLMs optimize each basic block separately. Note that we specifically use \textit{Instcombine} optimization which is applied to each basic block without additional transformation of CFG. However, removed instructions can have uses in other basic blocks. Thus, LLMs can produce instructions which arguments are initialized incorrectly. Nonetheless, we expect that LLMs should account for this (their advanced reasoning mechanisms should enable it) by defining the result registers for removed instructions when those registers might be used in other basic blocks. Note that the instructions without replacement can be removed only in the terminal CFG nodes, i.e. containing \textit{ret} as in the example from Table \ref{tab:code-comparison}.

\textbf{Register allocation.} Since LLMs only see the instructions within an optimized basic block, they can use only the registers that are used within that block. This is due to the fact that LLMs do not have the information about liveliness of all the registers.   

\textbf{The entire program representation.} Overall, we admit that that there could be also other limitations due to the fact that LLMs do not see the entire program code. Moreover, in certain cases, runtime support may be required, similar to what is implemented in binary translators.
However, we note that the main goal of our study was to investigate whether LLMs have sufficient capabilities to perform a simple peephole optimization and whether enhanced reasoning mechanisms can improve the models' performance.

In our future research, we will investigate the optimization capabilities of LLMs using a full program representation.

\section{Related work}
\label{sec:related_work}
Large Language Models (LLMs) have shown their efficiency in performing various tasks ranging from high-level code generation to compiler optimization and advanced reasoning.

\textbf{Large Language Models for Code Generation}.
In recent years, there has been a growing interest in leveraging Large Language Models (LLMs) for tasks involving source code generation. Models such as Copilot \cite {peng2023impactaideveloperproductivity}, Codex \cite {chen2021evaluatinglargelanguagemodels}, TransCoder \cite{lachaux2020unsupervisedtranslationprogramminglanguages}, CodeBERT \cite{feng2020codebertpretrainedmodelprogramming}, Code Llama \cite{rozière2024codellamaopenfoundation}, StarCoder \cite{li2023starcodersourceyou,lozhkov2024starcoder2stackv2}, Magicoder \cite{wei2024magicoderempoweringcodegeneration} and DeepSeek-Coder \cite{guo2024deepseekcoderlargelanguagemodel} have significantly advanced this field. These models support developers with tasks like code completion, generation, and translation across multiple programming languages. Open-sourced models like Code Llama and StarCoder have further empowered the community to adapt and fine-tune LLMs for specific software engineering needs.

\textbf{Large Language Models for Compilers}.
While LLMs have demonstrated significant potential in high-level code tasks, fewer models operate at the compiler level, particularly with code generation and compiler optimization. Most recent studies have focused on traditional machine learning methods for compiler optimization \cite{trofin2021mlgomachinelearningguided,wang2018machinelearningcompileroptimisation,Leather2020MachineLI,liang2023learningcompilerpassorders,DBLP:conf/mlsys/Haj-AliHMXAWS20,usingMachinelearnintofocusiterativeoptimization,10.1145/3427081.3427089}. Neural machine tranlation techiques have been employed to transform code between different representations, previous examples include compiling C to X86 assembly \cite{armengolestapé2022learningcx86translation} and decompiling assembly language to C \cite{armengolestapé2024sladeportablesmalllanguage,Hosseini_2022}. These works utilized smaller models or other deep learning methods. 
There are a few works related to using LLM at the compiler level. Examples include using large models for decompilers \cite{tan2024llm4decompiledecompilingbinarycode,wong2023refiningdecompiledccode,she2024wadecdecompilingwebassemblyusing}, LLVM-IR passes prediction with IR optimization \cite{cummins2023largelanguagemodelscompiler}, compiler fuzzing tests \cite{deng2023largelanguagemodelszeroshot,Yang_2024}.

\textbf{Recent Developments in Reasoning}. 
Enhancing the reasoning capabilities of LLMs has been a focus of recent research, particularly through methods like chain-of-thought (CoT) prompting. CoT encourages models to generate intermediate reasoning steps, improving performance on complex tasks such as mathematical reasoning and question answering \cite{NEURIPS2022_9d560961}. Chen \cite{li2024chaincodereasoninglanguage} proposed the chain-of-code (CoC) method. This approach enables models to handle tasks that require both logical and semantic reasoning, broadening the scope of questions LLMs can answer by "thinking in code". Experiments demonstrated that chain-of-code outperforms chain-of-thought and other baselines across various benchmarks. Gu et al. \cite{gu2024cruxevalbenchmarkcodereasoning} introduced CRUXEval (Code Reasoning, Understanding, and Execution Evaluation), a benchmark specifically designed to test input-output reasoning in code. Wang \cite{wang2024openropensourceframework} introduced OpenR, an open-source framework designed to enhance the reasoning capabilities of LLMs through reinforcement learning and process supervision. OpenR achieves advanced reasoning capabilities beyond traditional auto-regressive methods. To the best of our knowledge, there are no research studies which specifically investigate the enhanced reasoning mechanisms of LLMs for compiler optimizations.

\section{Conclusion}
\label{sec:conclusion}
In this paper, we investigate whether conventional LLMs, such as Llama2, can perform basic peephole compiler optimization, aiming to understand their performance and the errors they produce. Our findings suggest that one of the reasons for these errors is absence of enhanced reasoning capabilities. To explore this, we tested most advanced AI models from, OpenAI GPT-4o and GPT-o1, that implement such capabilities. Our results show that GPT-o1, despite not being fine-tuned, significantly outperforms both fine-tuned Llama2 and GPT-4o. We attribute this improvement to the chain-of-thought reasoning implemented in GPT-o1. We hope our work will inspire further research on using LLMs with enhanced reasoning mechanisms and chain-of-thought for code generation and optimization.

\bibliographystyle{IEEEtran}
\bibliography{references}
\onecolumn
\appendix
\section{Appendix: Examples of Optimizations}
Table \ref{tab:optimization_exmpl} provides examples of typical optimizations performed by peephole.

\begin{table*}[h]
\renewcommand{\arraystretch}{1.2}
\caption{Examples of typical optimizations performed by peephole.}
\label{tab:optimization_exmpl}
\centering
\small
\begin{tabular}{p{0.2\textwidth} p{0.25\textwidth} p{0.2\textwidth} p{0.2\textwidth}}
\toprule
\textbf{Optimization Type} & \textbf{Original Code} & \textbf{Optimized Code} & \textbf{Description} \\
\midrule

\textbf{Constant Folding} & 
\begin{minipage}[t]{\linewidth}
\begin{verbatim}
mov w0, #2
add w0, w0, #3
ret
\end{verbatim}
\end{minipage}
&
\begin{minipage}[t]{\linewidth}
\begin{verbatim}
mov w0, #5
ret
\end{verbatim}
\end{minipage}
&
The addition operation is folded into a single assignment. \\

\midrule
\textbf{Strength Reduction} & 
\begin{minipage}[t]{\linewidth}
\begin{verbatim}
mov w1, w0
mul w0, w1, #2
ret
\end{verbatim}
\end{minipage}
&
\begin{minipage}[t]{\linewidth}
\begin{verbatim}
lsl w0, w0, #1
ret
\end{verbatim}
\end{minipage}
&
Multiplication by 2 is reduced to a left shift operation. \\

\midrule
\textbf{Null Sequences} &
\begin{minipage}[t]{\linewidth}
\begin{verbatim}
lsl w8, w8, #1
lsr w8, w8, #0
ret
\end{verbatim}
\end{minipage}
&
\begin{minipage}[t]{\linewidth}
\begin{verbatim}
lsl w0, w0, #1
ret
\end{verbatim}
\end{minipage}
&
Right shift by 0 is effectively a no-op. \\

\midrule
\textbf{Combine Operations} &
\begin{minipage}[t]{\linewidth}
\begin{verbatim}
lsl x2, x1, #1
add x3, x2, x2
\end{verbatim}
\end{minipage}
&
\begin{minipage}[t]{\linewidth}
\begin{verbatim}
add x3, x1, x1
ret
\end{verbatim}
\end{minipage}
&
Two expressions are combined into one. \\

\midrule
\textbf{Algebraic Laws} &
\begin{minipage}[t]{\linewidth}
\begin{verbatim}
mov w9, wzr
mul w8, w8, w9
ret
\end{verbatim}
\end{minipage}
&
\begin{minipage}[t]{\linewidth}
\begin{verbatim}
mov w0, wzr
ret
\end{verbatim}
\end{minipage}
&
Any number multiplied by 0 is 0. \\

\midrule
\textbf{Address Mode Operations} &
\begin{minipage}[t]{\linewidth}
\begin{verbatim}
sub sp, sp, #16
.cfi_def_cfa_offset 16
mov w8, #1
str w8, [sp, #12]
ldr w0, [sp, #12]
add sp, sp, #16
ret
\end{verbatim}
\end{minipage}
&
\begin{minipage}[t]{\linewidth}
\begin{verbatim}
mov w0, #1
ret
\end{verbatim}
\end{minipage}
&
Optimizing address mode operations. \\
\bottomrule
\end{tabular}
\end{table*}
\end{document}